\documentclass[journal]{IEEEtran}
\ifCLASSINFOpdf
\else
\fi
\usepackage{epsfig}
\usepackage{graphicx}
\usepackage{amsmath}
\usepackage{amssymb}

\usepackage{algorithm}
\usepackage{algorithmic}
\usepackage{epsfig}
\usepackage{mathrsfs}
\usepackage{bm}
\usepackage{multirow}
\usepackage{color}
\usepackage[normalem]{ulem}
\usepackage{booktabs}

\newcommand{\bx}        {\textbf{x}}

\newcommand{\bD}        {\textbf{D}}
\newcommand{\bV}        {\textbf{V}}

\newcommand{\bY}        {\textbf{Y}}
\newcommand{\bM}        {\textbf{M}}

\newcommand{\btheta}    {\bm{\theta}}

\begin{document}
%
\title{Continual Few-Shot Learning with Adversarial Class Storage}
%
%
%


\author{Kun Wu,
        Chengxiang~Yin,~\IEEEmembership{Student Member,~IEEE},
        Jian~Tang,~\IEEEmembership{Fellow,~IEEE},
        Zhiyuan~Xu,~\IEEEmembership{Student Member,~IEEE},
        Yanzhi~Wang,~\IEEEmembership{Senior Member,~IEEE}
        and Dejun~Yang,~\IEEEmembership{Senior Member,~IEEE}
        
\thanks{Kun Wu, Chengxiang Yin, Jian Tang, and Zhiyuan Xu are with Department of Electrical Engineering and Computer Science,
Syracuse University, Syracuse, NY, 13244. E-mail: \{kwu102, cyin02, jtang02, zxu105\}@syr.edu.}
\thanks{Yanzhi Wang is with Department of Electrical and Engineering, Northeastern University, Boston, MA, 02115. E-mail: yanzhiwang@northeastern.edu.}
\thanks{Dejun Yang is with the Department of Computer Science,
Colorado School of Mines, Golden, CO 80401 USA. Email: djyang@mines.edu.}
}

\maketitle

\begin{abstract}
Humans have a remarkable ability to quickly and effectively learn new concepts in a continuous manner without forgetting old knowledge.
Though deep learning has made tremendous successes on various computer vision tasks,
it faces challenges for achieving such human-level intelligence.
In this paper, we define a new problem called continual few-shot learning,
in which tasks arrive sequentially and each task is associated with a few training samples.
We propose Continual Meta-Learner (CML) to solve this problem. 
CML integrates metric-based classification and a memory-based mechanism along with adversarial learning into a meta-learning framework,
which leads to the desirable properties:
1) it can quickly and effectively learn to handle a new task;
2) it overcomes catastrophic forgetting;
3) it is model-agnostic. We conduct extensive experiments on two image datasets, MiniImageNet and CIFAR100.
Experimental results show that CML delivers state-of-the-art performance in terms of classification accuracy on few-shot learning tasks without catastrophic forgetting.

\end{abstract} 



%
\IEEEpeerreviewmaketitle

\section{Introduction}
\IEEEPARstart{H}{umans} have a remarkable ability to effectively and quickly learn new concepts
in a continuous manner without forgetting old knowledge.
Even though deep learning 
has made tremendous successes on various computer vision tasks,
%
%
a huge amount of labeled data are usually needed to train a deep model,
which can then be used only for a specific task (e.g., classifying several types of animals).
Moreover, it has been reported that deep models may suffer from \textit{catastrophic forgetting}~\cite{Goodfellow2013}.
Hence, it is essential to improve deep learning such that a deep model can learn to handle a new task
from very limited training data without forgetting old knowledge.

To tackle the few-shot learning problem~\cite{Li2006,Lake2011},
a popular framework named meta-learning~\cite{Thrun1998} 
is introduced, in which
a meta-learner is required to guide a learner which can rapidly learn new concepts from a small dataset with only a few samples (e.g., 5) for each class.
Recently, a few methods~\cite{Finn2017,Snell2017,Gidaris2018}
have been proposed to solve the few-shot learning problem. Even though they can significantly reduce the amount of labeled
training data and help AI go one step closer to human-level intelligence, they may suffer from catastrophic forgetting, i.e.,
forgetting or even completely forgetting what they have already learned.

Continual learning focuses on how to achieve a good trade-off between learning new concepts and retaining old knowledge over a long time, which is known as the \textit{stability-plasticity dilemma}~\cite{Mermillod2013}.
Several regularization-based methods~\cite{Aljundi2018,Kirkpatrick2017,Li2018,Zenke2017} have been proposed for continual learning to alleviate catastrophic forgetting. 
Another approach~\cite{Draelos2017,Rusu2016} introduces additional neural resources to prevent catastrophic forgetting, but usually suffers from high complexities and high memory consumption. Moreover, none of these continual learning methods
address the issue of learning new concepts with very limited labeled data, which is a key challenge for achieving human-level AI.

Although solving the above problems is extremely significant, each of them suffers one-sidedness when depicting human-level learning.
%
%
In this paper, by introducing continual few-shot learning which is more realistic, we aim to bridge the gap between few-shot learning and continual learning by seeking a meta-learner, which can keep learning new concepts effectively and quickly from limited labeled data without forgetting old knowledge.
We consider a scenario, where during the meta-testing phase~\cite{Ravi2017}, new tasks arrive in a sequence rather than in a batch. In this case, a meta-learner has to learn for new tasks as regular meta-learners, and meanwhile, overcome catastrophic forgetting.

Even though an existing meta-learner (such as MAML~\cite{Finn2017}) can be slightly modified to deal with continual few-shot learning tasks,
we show via experiments that it suffers from catastrophic forgetting.
%
%
A straightforward solution to continual few-shot learning is to combine an existing meta-learner (such as MAML~\cite{Finn2017}) with a continual learning
method (such as MAS~\cite{Aljundi2018}). Unfortunately, our experimental results show that it still cannot effectively overcome the forgetting problem.
We propose a novel model-agnostic meta-learner, Continual Meta-Learner (CML), which
integrates metric-based classification and a memory-based mechanism along with adversarial
learning into an optimization-based meta-learning framework.
%
%
Comprehensive experiments have been conducted
on two widely-used datasets, MiniImageNet~\cite{Vinyals2016} and CIFAR100~\cite{Krizhevsky2009}.
The extensive experimental results have shown that CML delivers the state-of-the-art performance in terms of classification accuracy on few-shot learning tasks without catastrophic forgetting; 
while the optimization-based meta-learning approach (such as MAML) suffers from serious forgetfulness.
%
%
%

\begin{figure*}[!ht]
\setlength{\abovecaptionskip}{0pt}
\begin{center}
\includegraphics[scale=0.6]{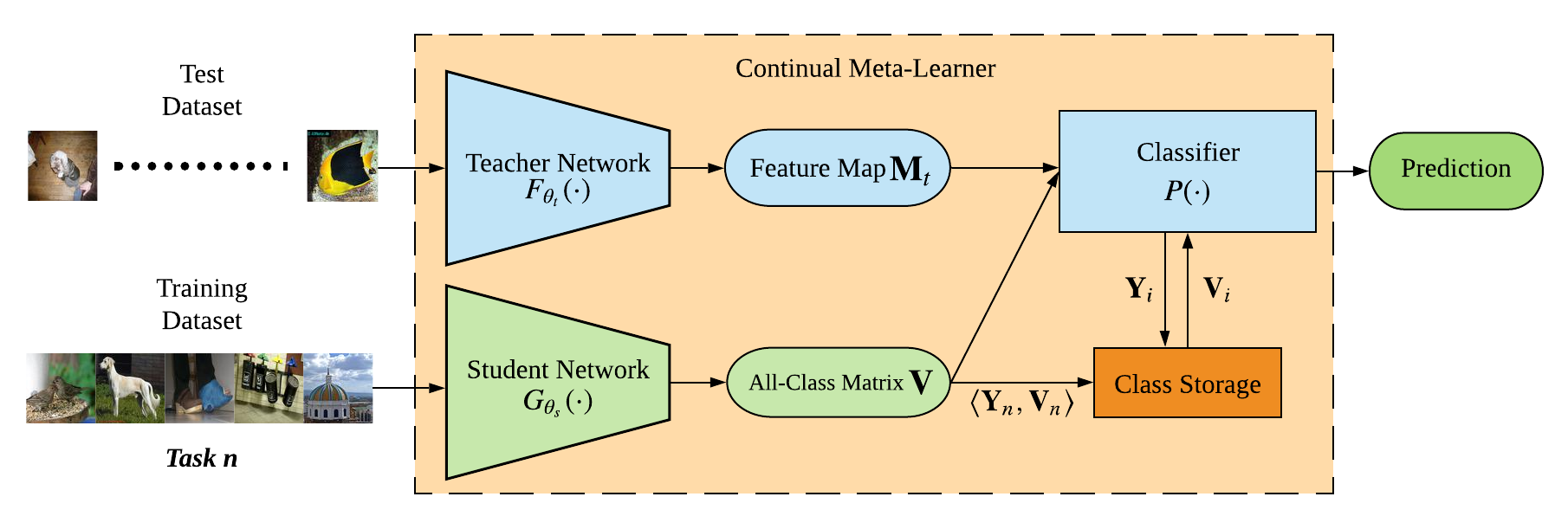}
\end{center}
\caption{Inference on the $n^{th}$ task of Continual Meta-Learner (CML). The Student Network takes $\bD_{train}$ to generate the all-class matrix $\bV_n$ for current task and store them into the Class Storage Module (CSM).
The pre-trained Teacher Network extracts the feature map $\bM_t$ for each test image of tasks $\mathcal{T}_{i} (i = 1, \dots, n)$.
The Classifier retrieves the all-class matrix $\bV_i$ for previous tasks $\mathcal{T}_{i} (i = 1, \dots, n-1)$ using the
label space $\bY_i$. 
Then the prediction is finished by the Classifier given the $\bM_t$ and corresponding $\bV_i (i = 1, \dots, n)$.}
\label{Fig:CML}
\end{figure*}

\section{Related Work}

\subsection{Few-Shot Learning}


There are a few methods exploiting similarity between samples of different classes for few-shot learning,
%
which can thus be called metric-based methods.
%
%
For example,
\cite{Vinyals2016} proposed Matching Networks,
which leverages the cosine similarity along with an attention mechanism for classification.
\cite{Snell2017} presented a model called Prototypical Networks for few-shot learning,
which is similar to Matching Networks, but leverages the Euclidean distance as the similarity metric.
%
\cite{Wang2018} proposed Prototype Matching Network, which combines the benefits of Prototypical networks and Matching networks. 
%
%
%
Other metric-based methods have been presented in~\cite{Gidaris2018,Oreshkin2018,Qi2018,Qiao2018,Sung2018}.
%

Another popular approach for few-shot learning focuses on optimizing model parameters, which can thus be called
the optimization-based approach.
%
%
Model-Agnostic Meta-Learner (MAML)~\cite{Finn2017} is a representative optimization-based method following the meta-learning framework, which
learns a good initialization for model parameters such that the model can be quickly
updated in several gradient descent steps to handle a new task. 
%
%
%
%
%
%
In a seminal work~\cite{Andrychowicz2016},
the authors showed the design of an optimization algorithm can be considered as a learning problem,
and designed a LSTM-based meta-learner. 
%
%
\cite{Ravi2017} proposed another LSTM-based meta-learner, which generates parameters sequentially with the objective of learning a good initialization.
Similarly,
\cite{Santoro2016} proposed a meta-learner based on a memory-augmented neural network, which leverages a high-capacity external
memory for few-shot learning.
%
%
%

These related works target only at few-shot learning
without addressing continual learning or the catastrophic forgetting issue, which, however, is the main focus of this work.

\subsection{Continual Learning}

Continual learning represents a long-standing challenge in the machine learning field~\cite{Hassabis2017}.
Several regularization-based methods 
have been proposed for continual learning, which impose regularization terms to restrain the update of model parameters.
For example,
\cite{Li2018} proposed a method called Learning without Forgetting (LwF), which uses knowledge distillation to
preserving seen knowledge.
%
\cite{Kirkpatrick2017} proposed Elastic Weight Consolidation (EWC), which
uses a quadratic penalty calculated by the difference between parameters for the new task and former tasks.
%
\cite{Zenke2017} leveraged intelligent synapses for continual learning.
%
%
\cite{Aljundi2018} presented Memory Aware Synapses (MAS), which calculates the importance of each parameter and penalizes
the change of important parameters when learning a new task to prevent forgetting.
\cite{Riemer2018} proposed Meta-Experience Replay (MER), which incorporates optimization based meta-learning method into experience play. 
Particularly, though they dealt with the continual learning problem through 
meta-learning, they didn't consider the few-shot setting as we considered in this paper.
%
%
%
Another widely-used approach is to leverage novel neural resources or extra memory to accommodate new concepts and retain old knowledge.
For exmaple,
\cite{Rusu2016} allocated new sub-networks for each new task while freezing the network trained from previous tasks.
Other works along this line include \cite{Fernando2017,Draelos2017}.
%
%
%
%
%
%
%

These related works presented solutions to the continual learning problem, whose setting is quite different from that of the
few-shot learning
problem. Hence, none of them can be directly applied to
few-shot learning
tasks.

\section{Problem Statement}
\label{Sec:Problem}

In the regular machine learning setting for classification, we aim to train a model representing a function $F(\cdot)$
which maps the given samples $\bx$ to the output $y$. Usually we train the parameters $\btheta$ of the model
on $\bD_{train}$ and evaluate it on $\bD_{test}$. While in the few-shot learning setting, we are interested
in meta-sets $\mathscr{D}$ which contain multiple regular datasets, and each dataset $\bD \in \mathscr{D}$
is split into $\bD_{train}$ and $\bD_{test}$ as those in regular machine learning.


In the few-shot learning problem,
we have meta-sets $\mathscr{D}_{meta-train}$ and $\mathscr{D}_{meta-test}$ used during
\emph{meta-training} and \emph{meta-testing} phases respectively~\cite{Ravi2017}.
It is important that the class labels in $\mathscr{D}_{meta-train}$ are not overlapping
with those of $\mathscr{D}_{meta-test}$.
%
%
Considering $N$-way, $K$-shot classification, for each task $\mathcal{T}_{i}$ with dataset $\bD \in \mathscr{D}$,  
$\bD_{train}$ contains $K$ samples
for each of $N$ classes, and $\bD_{test}$ contains samples for evaluation.
During the meta-training phase, we aim to train a meta-learner on $\mathscr{D}_{meta-train}$, 
which takes $\bD$ as input and then
produces a learner (with parameters $\btheta$) to maximize its classification accuracy
on the corresponding $\bD_{test}$. 
%
In contrast to regular machine learning, $\bD_{test}$ is used in the optimization.
We evaluate the performance of meta-learner on $\mathscr{D}_{meta-test}$ during the meta-testing phase.

In the continual learning setting, different tasks arrive in a sequence $[\mathcal{T}_1 \dots \mathcal{T}_n]$.
%
%
The most challenging part is that when training a new task $\mathcal{T}_j$, only $\bD_{train}$ of $\mathcal{T}_j$
is accessible, while $\bD_{train}$ of any previous task $\mathcal{T}_i$ $(i < j)$ is unavailable.

Here, we study a new continual few-shot learning problem,
%
where we have the meta-sets
consisting of $\mathscr{D}_{meta-train}$ and $\mathscr{D}_{meta-test}$,
which are the same as those of the few-shot learning problem.
%
Similarly, the learning can be divided into the meta-training and meta-testing;
however, during the meta-testing, 
instead of arriving in a batch like the few-shot learning, tasks arrive sequentially.
Moreover, in contrast to the continual learning,
each new task $\mathcal{T}_i$ only has a few samples (e.g., 5) in $\bD_{train}$ for each class.
Besides, we have prior knowledge (i.e., $\mathscr{D}_{meta-train}$), which can contribute to the meta-testing.
%
%
We believe that this problem setting is more realistic and can better represent the process
of continuously learning new concepts of a human.  The main challenge is to ensure that the learner can quickly, effectively and continuously learn new concepts without forgetting what it has already learned.
%


\section{Continual Meta-Learner (CML)}
\label{Sec:Method}

\subsection{Overview}
%
%
As illustrated in Figure~\ref{Fig:CML}, during the meta-testing phase, tasks arrive one by one in a sequence.
Every time when a new task $\mathcal{T}_i$ arrives, CML uses images from $\bD_{train}$ to quickly learn to handle the new task
and then takes images from $\bD_{test}$ as input and outputs the corresponding class labels.
CML consists of the following core components:

1) Teacher Network $F_{\btheta_t}(\cdot)$: It takes an image $\bx$ as input and extracts its features to form a feature map $\bM_t$ with a dimension $z$.

2) Student Network $G_{\btheta_s}(\cdot)$: 
As the core part of CML, it takes $\bD_{train}$ as input
and generates an \emph{all-class} matrix $\bV$.
Each row of $\bV$ corresponds to a class vector $\bV^l$ with dimension $z$,
which can be considered as 
a representation of $l^{th}$ class.

3) Classifier $P(\cdot)$: It takes the image feature map $\bM_t$ and the all-class matrix $\bV$ as input and predicts the class of image.
%

%
4) Class Storage Module (CSM): It stores $\bV$ of all previous tasks in a pair form of label and class vector $\langle y^l, \bV^l\rangle$, which are used to overcome catastrophic forgetting.

In addition, we introduce a Discriminator $D_{\btheta_d}(\cdot)$: 
It tries to distinguish between the image feature map $\bM_t$ (belonging to class $l$) 
and the corresponding class vector $\bV^l$ during the meta-training phase. 
It is trained with the Student Network in an adversarial manner and can guide the Student Network for better class vectors. 

When a new task $\mathcal{T}_i$ arrives during the meta-testing phase,
CML takes $\bD_{train}$ as input, and performs \textit{fast-learning}
%
%
to generate the all-class matrix $\bV$ 
and 
stores $\bV$ in Class Storage.
When evaluating with $\bD_{test}$, the Teacher Network generates the feature map $\bM_t$ for each input image $\bx$,
and passes $\bM_t$ to the Classifier
%
to predict the class of image
by calculating the similarity between 
$\bM_t$ and
each class vector $\bV^{l}$ in $\bV$ learned from the fast-learning.
%
The design of CML leads to the following desirable properties:

1) \textit{Learning quickly and effectively from a few samples}: CML learns a good initialization of the Student Network
after the meta-training phase such that it can adapt to any new task with only a few samples. Moreover,
the Student Network and the Discriminator are trained in an adversarial manner such that the generated class vectors
are close to the best ones, which is the key to accurate classification.

2) \textit{Overcoming catastrophic forgetting}: Even though the new tasks come in a continual manner, CML learns to handle each new task independently with the same good initialization (i.e., $\btheta$),
which would not reduce the performance of the former tasks or interfere with the learning for new tasks.
To conquer the catastrophic forgetting, CML stores the all-class matrix $\bV$ in the Class Storage Module (CSM), which can be retrieved when needed, i.e., will not be forgotten.
Moreover, it only needs a small memory footprint.

3) \textit{Model-agnostic}: As MAML, CML can be applied to any model that can be trained by gradient descent.


\begin{algorithm}[t]
\caption{Meta-training Algorithm for CML}
\label{Alg:CML}

{\bf Require:} The pre-trained Teacher Network $F_{\btheta_t}(\cdot)$ \\
{\bf Require:} Task distribution $p(\mathcal{T})$ \\
{\bf Require:} $\alpha_s/\alpha_d, \beta_s/\beta_d$: the step sizes for fast-learning and meta-update respectively
\begin{algorithmic}[1]
\STATE Randomly initialize $\bm\btheta_s, \bm\btheta_d$;
\WHILE{not done}
\STATE Sample a batch of tasks $\mathcal{T}_i \sim p(\mathcal{T})$;

\FORALL{$\mathcal{T}_i$}
\STATE Obtain $\bD_{train}$ from $\mathcal{T}_i$;
\STATE all-class matrix $\bV_i \gets G_{\btheta_s}(\bx)$ where $\bx \in \bD_{train};$
\STATE $\mathcal{L}_{train}^s (\btheta_s) \gets \frac{1}{|\bD_{train}|} \sum\limits_{(\bx, y) \in \bD_{train}} \ell_{s}(\btheta_s; \bx, y, \bV_i)$;
\STATE $\btheta_{i,s}^{'} \gets \btheta_s - \alpha_s \nabla_{\btheta_s} \mathcal{L}_{train}^s(\btheta_s)$;
\STATE $\mathcal{L}_{train}^d(\btheta_d) \gets \frac{1}{|\bD_{train}|} \sum\limits_{\bx \in \bD_{train}} \ell_d(\btheta_d;\bx)$;
\STATE $\btheta_{i,d}^{'} \gets \btheta_d - \alpha_d \nabla_{\btheta_d} \mathcal{L}_{train}^d (\btheta_d)$;

\STATE all-class matrix $\bV_{i}^{'} \gets G_{\btheta_{i,s}^{'}}(\bx)$ where $\bx \in \bD_{train};$

\ENDFOR

\FORALL{$\mathcal{T}_i$}
\STATE Obtain $\bD_{test}$ from $\mathcal{T}_i$;
\STATE{$\mathcal{L}_{test}^s(\btheta_{i,s}^{'}) \gets \frac{1}{|\bD_{test}|} \sum\limits_{(\bx, y) \in \bD_{test}} (\ell_{s}(\btheta'_{i,s}; \bx, y, \bV_{i}^{'}) + \ell_{d}^-(\btheta'_{i,s};\bx));$}

\STATE $\mathcal{L}_{test}^d(\btheta_{i,d}^{'}) \gets \frac{1}{|\bD_{test}|} \sum\limits_{\bx \in \bD_{test}} \ell_d(\btheta_{i,d}^{'};\bx)$;

\ENDFOR
\STATE $\btheta_s \gets \btheta_s - \beta_s \nabla_{\btheta_s} \sum_{\mathcal{T}_i} \mathcal{L}_{test}^s(\btheta_{i,s}^{'})$;

\STATE $\btheta_d \gets \btheta_d - \beta_d \nabla_{\btheta_d} \sum_{\mathcal{T}_i} \mathcal{L}_{test}^d(\btheta_{i,d}^{'})$;

\ENDWHILE

\end{algorithmic}
\end{algorithm}

\subsection{Meta-training}
We formally present the proposed CML as Algorithm \ref{Alg:CML} for \textit{meta-training}.
%
In the beginning, we pre-train a teacher network
on the meta-set $\mathscr{D}_{meta-train}$ such that it gains enough knowledge to serve
as a ``teacher".
%
%
After pre-training, we discard the last fully-connected layer and
keep the the remaining network
as the Teacher Network, whose parameters $\btheta_t$ will be fixed during the meta-training and the meta-testing phases. 

During the meta-training phase, CML takes a batch of tasks $\{\mathcal{T}_1, \cdots, \mathcal{T}_n\}$ 
($n=4$ in our implementation) as input
and learns a good initialization of the Student Network and the Discriminator.
The entire meta-training consists of two phases: \textit{fast-learning} and \textit{meta-update}. \\
%
%

\noindent \textbf{Fast-learning} \qquad
CML learns from $\bD_{train}$ of each
individual task in the batch (Line 4-12).
Given an input image $\bx$ ($84 \times 84$ in our implementation),
the Teacher Network
yields a feature map $\bM_t$ with a dimension $z$, and the Student Network yields a feature map $\bM_s$ with the same dimension.
%
For the 1-shot setting, we directly obtain the class vector $\bV^l$ for the input image belongs to the $l^{th}$ class, i.e., $\bV^l=\bM_s$.
%
For the K-shot setting,
we obtain the class vector $\bV^l$ by taking the mean values.
%
%
The all-class matrix $\bV$ is then constructed by stacking the class vectors all together.
$\bV$ has a dimension of $N \times z$, where $N$ is the number of classes ($N=5$ in our implementation) and $z$ is the dimension of each class vector $\bV^l$ ($z=512$ in our implementation).
%
%
%
%

Then Classifier calculates the cosine similarity between 
the feature map $\bM_t$
and each class vector $\bV^l$ 
for the prediction score of each class:
%
\begin{align}
\label{Equatio: classifier}
    P(\bM_t, \bV) = \text{softmax}(Cos(\bM_t, \bV^{\top})),
\end{align}
where $Cos(\cdot)$ calculates the cosine similarity.
We choose the cosine similarity because it eliminates the interference
resulting from different orders of magnitude corresponding to different classes.
%
%
After that, we use a softmax function to normalize prediction scores.

To train the Student Network during the fast-learning, we use the cross-entropy loss $\ell_{s}$ on the Classifier:
%
\begin{align}
\label{Equation: ell_sp}
    \begin{split}
    \ell_{s}(\btheta_s; \bx, y, \bV) = - [y \log P(F_{\btheta_t}(\bx), \bV) \\ 
    + (1-y) \log (1-P(F_{\btheta_t}(\bx), \bV))],
    \end{split}
\end{align}
where $(\bx, y)$ is an image/label pair from $\bD_{train}$ 
and $\bV$ is the all-class matrix.


%
%
%

The Student Network is trained by minimizing the loss $\ell_{s}$ through gradient descent (Line 6-8).
%
Note that it is trained for each task $\mathcal{T}_i$ independently but
with the same parameters $\btheta_s$.
%
Through fast-learning, $\btheta_{i,s}^{'}$ is calculated for each task $\mathcal{T}_i$ based on $\btheta_s$.
%
%
%

The Discriminator takes the feature map $\bM_t=F_{\btheta_t}(\bx)$ from the Teacher Network
as the real (i.e., true) input and the feature map $\bM_s=G_{\btheta_s}(\bx)$ from the Student Network as the fake (i.e., false) input. Its loss $\ell_d$ is:
\begin{align}
\label{Equation: ell_d}
    \ell_d(\btheta_d;\bx) = - (\ell_d^+(\btheta_d;\bx) + \ell_d^-(\btheta_d;\bx))
\end{align}
where 
\begin{align}
    \label{Equation: ell_d_pos}
    \ell_d^+(\btheta_d;\bx) &= \log D_{\btheta_d}(F_{\btheta_t}(\bx)),  \\
    \label{Equation: ell_d_neg}
    \ell_d^-(\btheta_d;\bx) &= \log (1-D_{\btheta_d}(G_{\btheta_s}(\bx))),
\end{align}
and $\bx$ is an image from $\bD_{train}$. 
%
Similar to the Student Network,
$\btheta_{i,d}^{'}$ is calculated based on $\btheta_d$ through fast-learning (Line 9-10).

%
The small amount of training data and gradient descent steps (e.g., 1 sample and 5 steps) lead to a limited learning ability of fast-learning.
In such kind of condition, during the fast-learning phase, the Student Network should devote into the classification task (i.e., $\ell_{s}$), rather than interfered by the adversarial training. 
As discussed above, the adversarial loss (i.e, $\ell_d$) is only used to train the Discriminator in fast-learning.

Through fast-learning phase, for each task, CML stores its all-class matrix $\bV$ in a pair form of label and class vector $\langle y^l, \bV^l\rangle$, which only consumes a very small space of 12KB. Hence CML has a low memory footprint.\\

\begin{table*}[htbp]\scriptsize
\centering
\caption{Classification accuracies for 4 tasks at the end of the sequence on MiniImageNet (5-way, 1-shot)}
\renewcommand\arraystretch{1.3}
\begin{tabular}{l|cccc|c}
\toprule
Method & $\mathcal{T}_1$ & $\mathcal{T}_2$ & $\mathcal{T}_3$ & $\mathcal{T}_4$ & Average \\ \midrule
MAML~\cite{Finn2017}+FT & $26.85\pm1.62\%$ & $27.84\pm1.60\%$ & $30.61\pm1.63\%$ & $34.42\pm1.50\%$ & $29.93\%$\\ 
MAML~\cite{Finn2017}+MAS~\cite{Aljundi2018} & $27.20\pm1.52\%$ & $27.99\pm1.61\%$ & $31.60\pm1.62\%$ & $34.45\pm1.60\%$ & $30.31\%$ \\ 
MN~\cite{Vinyals2016}+CSM & $43.01\pm1.14\%$ & $43.54\pm1.17\%$ & $43.18\pm1.18\%$ & $43.31\pm1.21\%$ & $43.26\%$ \\ 
PN~\cite{Snell2017}+CSM & $48.19\pm0.87\%$ & $48.63\pm0.91\%$ & $48.39\pm0.83\%$ & $48.57\pm0.88\%$ & $48.44\%$ \\ \cmidrule(lr){1-6}
CML(Ours) & \bm{$57.86\pm1.47\%$} & \bm{$60.70\pm1.42\%$} & \bm{$57.85\pm1.49\%$} & \bm{$58.23\pm1.38\%$} & \bm{$58.66\%$} \\ \bottomrule

\end{tabular}
\label{mini-1shot}
\end{table*}

\begin{table*}[htbp]\scriptsize
\centering
\caption{Classification accuracies for 4 tasks at the end of the sequence on MiniImageNet (5-way, 5-shot)}
\renewcommand\arraystretch{1.3}
\begin{tabular}{l|cccc|c}
\toprule
Method & $\mathcal{T}_1$ & $\mathcal{T}_2$ & $\mathcal{T}_3$ & $\mathcal{T}_4$ & Average \\ \midrule
MAML~\cite{Finn2017}+FT & $29.74\pm1.91\%$ & $29.02\pm1.76\%$ & $34.04\pm1.80\%$ & $49.34\pm1.60\%$ & $35.53\%$ \\
MAML~\cite{Finn2017}+MAS~\cite{Aljundi2018} & $28.97\pm1.71\%$ & $26.92\pm2.00\%$ & $35.09\pm1.90\%$ & $50.52\pm1.61\%$ & $35.37\%$ \\
MN~\cite{Vinyals2016}+CSM & $53.68\pm1.02\%$ & $54.11\pm0.98\%$ & $54.13\pm1.03\%$ & $53.77\pm1.02\%$ & $53.92\%$ \\
PN~\cite{Snell2017}+CSM & $66.08\pm0.72\%$ & $66.80\pm0.73\%$ & $66.16\pm0.71\%$ & $66.78\pm0.72\%$ & $66.45\%$ \\ \cmidrule(lr){1-6}
CML(Ours) & \bm{$70.86\pm1.68\%$} & \bm{$70.85\pm1.55\%$} & \bm{$71.09\pm1.65\%$} & \bm{$69.11\pm1.81\%$} & \bm{$70.48\%$} \\ \bottomrule

\end{tabular}
\label{mini-5shot}
\end{table*}

\noindent \textbf{Meta-update} \qquad
Both of the Student Network and the Discriminator are trained on $\bD_{test}$ across all tasks during meta-update (Line 18-19) to arrive a good initialization from where a new task can be handled with a few samples (e.g., 1 or 5).
%

For the Student Network, during meta-update, the cross-entropy loss $\ell_{s}$ (Equation~\ref{Equation: ell_sp}, Line 15) for classification is computed based on images $\bx$ from $\bD_{test}$ and the updated all-class matrix $\bV^{'}$ through the fast-learning.
In addition, the same adversarial loss $\ell_d(\btheta_d;\bx)^-$ as Equation~\ref{Equation: ell_d_neg} is added (Line 15).
%
%
%
%
%
%
Similar to the Generative Adversarial Networks (GANs)~\cite{Goodfellow2014}, 
the Student Network can
be considered a generator to generate the feature maps $\bM_s$.
Meanwhile, the Discriminator is used to distinguish between $\bM_s$ and $\bM_t$, which is generated by the Teacher Network.
%

For the meta-update of the Discriminator, we use the same binary-entropy loss $\ell_d$ (Equation~\ref{Equation: ell_d}, Line 16) given the images $\bx$ in $\bD_{test}$.

The meta-update of the proposed meta-training algorithm can 
lead to a more representative and discriminative all-class matrix $\bV$ during meta-testing,
%
which is contributed by two factors:
1) The Teacher Network, through pre-training on the meta-set $\mathscr{D}_{meta-train}$, can generate a good feature map $\bM_t$ for each image $\bx$, which serves as a learning template for the Student Network.
2) In contrast to $\bD_{train}$, the $\bD_{test}$ used in the meta-update consists of more samples for each class (e.g., 15), 
which does not have the limitation of few-shot setting in the fast-learning phase.
Thus the adversarial training, by adding the adversarial loss (i.e., $\ell_{d}^-$) during the meta-update, 
can guide the Student Network to achieve a better initialization from where a more discriminative matrix can be obtained.
%
%
As a matter of course, a better all-class matrix $\bV$ stands for higher classification accuracy for each new task during the meta-testing.
%
%
%
On the other hand, the cross-entropy loss $\ell_{s}$ for classification can stabilize the adversarial training of the Student Network and the Discriminator. 
%
Hence, in the meta-update, $\ell_{s}$ and $\ell_{d}^-$ can benefit from each other.

%
%
%
%
%
%
One-step gradient descent is performed on $\btheta_s$ and $\btheta_d$ to update the Student Network and the Discriminator respectively (Line 18-19).
Similar to the MAML~\cite{Finn2017}, 
meta-update is performed
on the parameters $\btheta_s$ and $\btheta_d$ rather than $\btheta_{i,s}^{'}$ and $\btheta_{i,d}^{'}$,
while the losses $\ell_{s}$, $\ell_{d}^-$ and $\ell_d$ are computed based on the calculated
parameters $\btheta_{i,s}^{'}$ and $\btheta_{i,d}^{'}$ 
through
fast-learning.
Note that the losses for meta-update are computed across all the tasks $\{\mathcal{T}_1, \cdots, \mathcal{T}_n\}$ in the batch rather than a single task.
In this way, CML can learn a good initialization of both the Student Network and the Discriminator such that
it can quickly learn to deal with a new task during the meta-testing phase.



\subsection{Meta-testing}
During the meta-testing phase,
we have $\btheta_{s}$ of the Student Network 
as a good initialization to handle the new tasks.
%
Note that the Discriminator is discarded, 
since it is only used to assist the training of the Student Network in the meta-update, rather than the fast learning, during the meta-training phase.

As mentioned in Section 3, for continual few-shot learning, tasks arrive sequentially during the meta-testing phase.
For the new task in sequence,
similar to the meta-training phase,
the Student Network has the same fast-learning process 
for quickly learning from the initialization $\btheta_{s}$ with a few samples. 
Through the fast-learning on $\bD_{train}$, the Student Network with the calculated parameters $\btheta_{s}^{'}$ generates an all-class matrix $\bV$ for the current task
and falls back to the initialization $\btheta_{s}$ for the next task. 
Then, for an image from $\bD_{test}$, with the feature map $\bM_t$ generated from the fixed Teacher Network,
the Classifier predicts which class the image belongs to, and stores the all-class matrix $\bV$ in a pair form of label and class vector $\langle y^l, \bV^l\rangle$ into the Class Storage Module (CSM).
%
When performing inference on a former task without accessible $\bD_{train}$ for the continual few-shot learning,
the Classifier retrieves the pairs $\langle y^l, \bV^l\rangle$ from the CSM according to the class labels of the former task and forms an all-class matrix $\bV$ for the task, which is used to predict the class of the images from $\bD_{test}$.

In this way, for each new task, the Student Network of CML always learns from the same initialization $\btheta_{s}$ to generate the all-class matrix $\bV$, which is stored independently and can be retrieved accordingly. Hence, the inference on the new task does not affect that on the former task, and the catastrophic forgetting is conquered at the cost of a small memory footprint.
%




\subsection{Implementation Details}
For the Teacher Network, we use ResNet18 following~\cite{He2016}. 
Given an input image $\bx$ with a size of $84 \times 84$, the ResNet18 yields a feature map $\bM$ with a dimension of $z=512 \times 1 \times 1 = 512$.

The Student Network contains a Convolutional Neural Network (CNN) similar to that in~\cite{Finn2017},
which has 4 convolutional modules, each of which contains a $3 \times 3$ convolution layer followed by batch normalization~\cite{Ioffe2015}, a ReLU nonlinearity and $2 \times 2$ max pooling.
It has 64 filters in the first two convolutional layers and 128 filters in the last two convolutional layers.
Given an input image $\bx$ with a size of $84 \times 84$,
the Student Network yields a feature map $\bM$ with a dimension of $z=128 \times 2 \times 2 = 512$, which is same as ResNet18.

We use a multilayer perceptron (MLP)~\cite{Goodfellow2016} to implement the Discriminator, which contains two fully-connected layers.
The first fully-connected layer is followed by batch normalization and a ReLU nonlinearity; and the second fully-connected layer is followed by
the sigmoid function that normalizes output.

In addition, in our implementation, during fast-learning, we set the number of training steps to 5, and the learning rates $\alpha_s$
and $\alpha_d$ of both the Student Network and the Discriminator to 0.01.
During meta-update, the learning rates $\beta_s$ and $\beta_d$ of the Student Network and the Discriminator are set to 0.1 and 0.001 respectively,
which both decay rapidly with the epoch.
\section{Performance Evaluation}
\label{Sec:Eval}

\begin{table*}[t]\scriptsize
\centering
\caption{Classification accuracies for 4 tasks at the end of the sequence on CIFAR100 (5-way, 1-shot)}
\renewcommand\arraystretch{1.3}
\begin{tabular}{l|cccc|c}
\toprule
Method & $\mathcal{T}_1$ & $\mathcal{T}_2$ & $\mathcal{T}_3$ & $\mathcal{T}_4$ & Average \\ \midrule
MAML~\cite{Finn2017}+FT & $31.03\pm1.73\%$ & $33.25\pm2.12\%$ & $34.07\pm1.79\%$ & $40.31\pm1.84\%$ & $34.66\%$ \\ 
MAML~\cite{Finn2017}+MAS~\cite{Aljundi2018} & $31.47\pm1.90\%$ & $32.84\pm1.88\%$ & $34.45\pm1.72\%$ & $40.69\pm1.89\%$ & $34.86\%$ \\ 
MN~\cite{Vinyals2016}+CSM & $52.93\pm1.13\%$ & $51.66\pm1.38\%$ & $51.90\pm1.40\%$ & $51.74\pm1.34\%$ & $52.05\%$ \\ 
PN~\cite{Snell2017}+CSM & $49.38\pm0.87\%$ & $49.38\pm0.85\%$ & $48.38\pm0.85\%$ & $49.24\pm0.86\%$ & $49.09\%$ \\ \cmidrule(lr){1-6}
CML(Ours) & \bm{$61.57\pm1.58\%$} & \bm{$62.80\pm1.49\%$} & \bm{$61.06\pm1.63\%$} & \bm{$62.91\pm1.56\%$} & \bm{$62.08\%$} \\ \bottomrule

\end{tabular}
\label{CIFAR-1shot}
\end{table*}

\begin{table*}[t]\scriptsize
\centering
\caption{Classification accuracies for 4 tasks at the end of the sequence on CIFAR100 (5-way, 5-shot)}
\renewcommand\arraystretch{1.3}
\begin{tabular}{l|cccc|c}
\toprule
Method & $\mathcal{T}_1$ & $\mathcal{T}_2$ & $\mathcal{T}_3$ & $\mathcal{T}_4$ & Average \\ \midrule
MAML~\cite{Finn2017}+FT & $29.24\pm2.01\%$ & $33.21\pm2.03\%$ & $40.34\pm1.76\%$ & $58.18\pm1.73\%$ & $40.24\%$ \\ 
MAML~\cite{Finn2017}+MAS~\cite{Aljundi2018} & $31.22\pm1.90\%$ & $33.48\pm1.95\%$ & $40.92\pm1.82\%$ & $57.42\pm1.68\%$ & $40.76\%$ \\ 
MN~\cite{Vinyals2016}+CSM & $66.27\pm1.13\%$ & $65.00\pm1.20\%$ & $65.69\pm1.06\%$ & $65.30\pm1.21\%$ & $65.56\%$ \\ 
PN~\cite{Snell2017}+CSM & $70.39\pm0.70\%$ & $70.03\pm0.70\%$ & $69.02\pm0.74\%$ & $69.64\pm0.68\%$ & $69.77\%$ \\ \cmidrule(lr){1-6}
CML(Ours) & \bm{$74.04\pm1.21\%$} & \bm{$74.91\pm1.17\%$} & \bm{$73.38\pm1.23\%$} & \bm{$73.79\pm1.33\%$} & \bm{$74.03\%$} \\ \bottomrule

\end{tabular}
\label{CIFAR-5shot}
\end{table*}

\subsection{Experimental Setup}
\label{Sec:Setup}

In our experiments, we evaluated the proposed CML on two widely-used image datatsets, MiniImageNet~\cite{Vinyals2016} and CIFAR100~\cite{Krizhevsky2009}.
MiniImageNet includes 100 classes with 600 samples per class and each image has a spatial size of $84 \times 84$.
CIFAR100 is another image dataset, which also has 100 classes and each class contains 600 samples too.
Both datasets were divided into three disjoint meta-sets with 64, 16, and 20 classes for meta-training, meta-validation and meta-testing respectively.
%
%
Following the settings used in~\cite{Finn2017}, each task was set to a $5$-way, $K$-shot ($K=1$ or $5$ in the experiments).
We firstly sampled $5$ classes from the meta-training dataset.
For each class, we sampled $K$ images for training and $15$ images for testing.
Then the training set $\bD_{train}$ includes $5K$ images and the testing set $\bD_{test}$ contains $15*5=75$ images.
%
%
During the meta-testing phase, tasks from the meta-testing set (with unseen classes) arrive one by one in a sequence.
%
%
Whenever training for a new task $\mathcal{T}_j$ was completed, we also evaluated each method over
all the previous tasks $\{\mathcal{T}_i\} (\forall i<j)$ using their $\bD_{test}$.
%

%
%

We are the first to consider the continual few-shot learning problem, which cannot be solved directly by any existing few-shot learning or
continual learning method.
For the purpose of performance evaluation, we chose several recent and well-known meta-learners~\cite{Finn2017,Snell2017,Vinyals2016}
as the baselines and slightly modified them to handle the continual few-shot learning problem studied here.
%
%
Specifically, MAML~\cite{Finn2017} was used as a baseline, which is a representative optimization-based meta-learner.
We only fine-tune the MAML model whenever a new task arrives and call it \textbf{MAML+FT}.
Moreover, for fair comparisons, we also extended MAML to support continual learning by integrating MAS~\cite{Aljundi2018} into its optimization framework, which
is a well-known regularization-based approach proposed particularly for continual learning.
Specifically, we computed the importance for each parameter in the model using the method in MAS,
added them as a regularization term into the loss function and meta-trained the model using MAML's optimization framework.
This method is called \textbf{MAML+MAS} in the following.
For both MAML+FT and MAML+MAS, since tasks arrive one by one in a sequence (instead of in a batch), we updated the models sequentially using data corresponding
to each incoming task.
In addition, Matching Networks~\cite{Vinyals2016} (MN) and Prototypical Networks~\cite{Snell2017} (PN) are both metric-based
few-shot learning
methods, which were used as baselines too.
Since they do not have the fast-learning phase as MAML and CML, which learns from the $\bD_{train}$ of each arriving task,
they can't be fine-tuned on current task.
Again for fair comparisons, during meta-testing, we upgraded both methods as \textbf{MN+CSM} and \textbf{PN+CSM} by integrating our Class Storage Module (CSM) and
storing the class prototypes for each task, 
which helps them
overcome catastrophic forgetting as CML.

\begin{figure*}[t]
\centering
\includegraphics[scale=0.57]{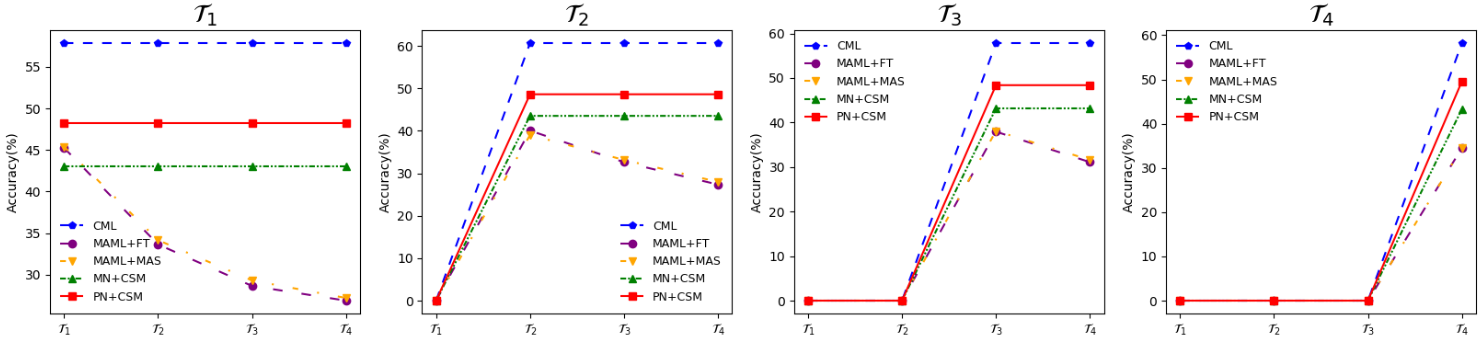}
\caption{Classification accuracies on the timeline of 4 tasks sequence on MiniImageNet (5-way, 1-shot)}
\label{Fig:MiniImageNet_1shot}
\end{figure*}


\begin{figure*}[t]
\centering
\includegraphics[scale=0.57]{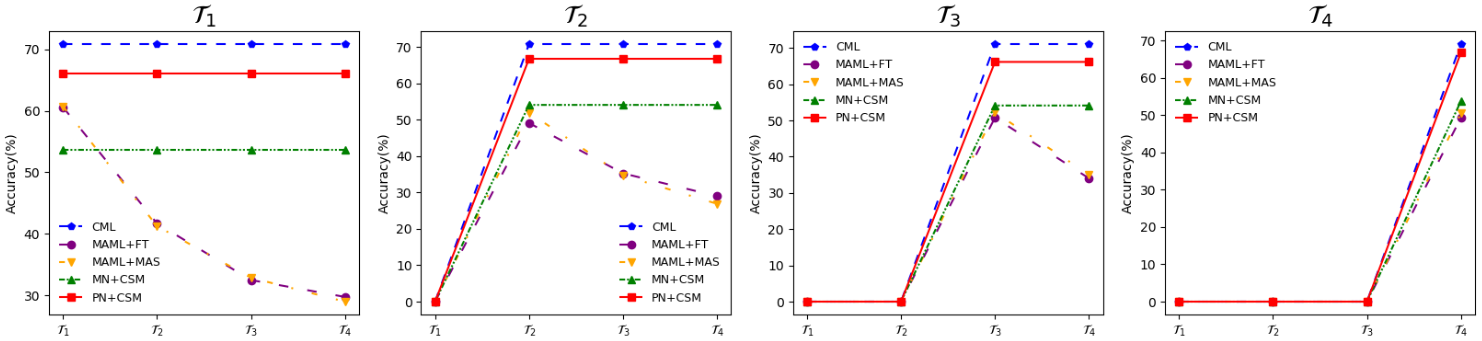}
\caption{Classification accuracies on the timeline of 4 tasks sequence on MiniImageNet (5-way, 5-shot)}
\label{Fig:MiniImageNet_5shot}
\end{figure*}

\subsection{Experimental Results}
\label{Sec:Results}

%
Tables~\ref{mini-1shot} and~\ref{mini-5shot} show the image classification accuracy (with a confidence interval of $95\%$) given by each method
on each task at the end of the 4-tasks sequence on the MiniImageNet dataset.
Tables~\ref{CIFAR-1shot} and~\ref{CIFAR-5shot} show the corresponding results on the CIFAR100 dataset.
We can make the following observations from these results.
%

First of all, CML consistently outperforms all the other methods by relatively large margins
on all the 4 tasks in both 5-way, 1-shot and 5-way, 5-shot cases on both datasets.
%
%
For instance, from Table~\ref{mini-1shot}, we can see that CML achieves an average accuracy of $58.66\%$,
while the average accuracies given by the other four baselines are
$48.44\%, 43.26\%, 30.31\%. 29.93\%$ respectively. Similar observations can be made from the other three tables.
This observation shows that the proposed CML is indeed effective for continual few-shot learning and is superior to
the other meta-learners in terms of classification accuracy.

%
%

%
Second, MAML+FT and MAML+MAS suffer from catastrophic forgetting.
For instance, from Table~\ref{mini-5shot}, we can see that at the end of the 4-tasks sequence,
MAML+MAS provides an accuracy of $50.52\%$ for task $\mathcal{T}_4$, while accuracies on
the previous tasks are pretty low; and the earlier the task, the worse the accuracy.
%
%
This observation shows that a regularization-based method is not effective for mitigating the forgetting problem
in the context of few-shot learning.
Even though MAML+MAS calculates the importance of each parameter in model
and uses them as a regularization term to penalize the change of model parameters,
the numbers of training samples ($N \times K$) and training steps (typically no more than 5)
are too small to precisely measure the importance of parameters and give proper penalties.
Hence, MAS may work well regular continual learning task but turns out to be ineffective for
continual few-shot learning. This also indicates that regularization may not be a right way for
dealing with continual few-shot learning tasks.

In contrast, CML successfully overcomes catastrophic forgetting by consistently maintaining high accuracies on both the most recent task
and the previous tasks. For example, from Table~\ref{mini-5shot}, we can see that CML achieves a high average accuracy
of $70.48\%$, and more importantly, the accuracies on the $4$ tasks are all close to $70\%$.
%
%
CML delivers such superior performance in this regard because it tackles the forgetting problem with
a rather direct mechanism by storing the class vectors in the memory with a small footprint and retrieving them
when needed.
%

Third, from these tables, we also notice that MN+CSM and PN+CSM well overcome catastrophic forgetting by maintaining high accuracies for all the previous tasks.
These results further show the effectiveness of the memory-based mechanism for continual learning, which has been utilized by CML, MN+CSM and PN+CSM.
%
%
Even though MN+CSM and PN+CSM have the ability to overcome catastrophic forgetting,
CML beats them by relatively large margins in terms of classification accuracies, especially on the 1-shot learning tasks. For example, from Table~\ref{mini-1shot},
CML improves the average accuracy from $43.26\%$ and $48.44\%$ to
$58.66\%$.
As mentioned above, 
CML integrates metric-based classification and adversarial learning
into an optimization-based meta-training framework which, we believe, leads to performance improvements
over the state-of-the-art metric-based methods on these few-shot learning tasks.

To better show the change of accuracy during the arriving process of tasks, we plot Figures~\ref{Fig:MiniImageNet_1shot} and ~\ref{Fig:MiniImageNet_5shot}.
%
In these figures, the x-axis gives the time-step corresponding to the completion of learning for each task and each sub-figure shows the change of accuracy of each task following the timeline. Note that at the time-step of task $\mathcal{T}_j$, we cannot measure the accuracy of task $\mathcal{T}_i (i>j)$, so the corresponding accuracies are simply set to $0$. For example, on the
third sub-figure (corresponding to task $\mathcal{T}_3$) of Figure~\ref{Fig:MiniImageNet_1shot}, we can see that the accuracies corresponding to the time-steps of tasks $\mathcal{T}_1$ and $\mathcal{T}_2$ are set to $0$. From this figure, we can clearly see that CML, MN+CSM and PN+CSM can well overcome catastrophic forgetting. For example, in the first sub-figure of Figure~\ref{Fig:MiniImageNet_1shot}, the curves corresponding to these three algorithms are basically straight horizontal lines, i.e., the
accuracies do not change with the arrival of new tasks.
However, both MAML+FT and MAML+MAS suffer from serious forgetting problems. For instance, in the same sub-figure, if MAML+FT is used, the accuracy of task $\mathcal{T}_1$ drops sharply from over $45\%$, which is higher than that given by MN+CSM,
all the way down to $26.85\%$ during the arriving process of all the 4 tasks. 
This observation further shows that MAML is indeed an effective meta-learner for few-shot learning setting 
but behaves poorly on continual learning due to lack of an effective mechanism for retaining old knowledge.
%
%
Due to the limited space, we cannot include the figures corresponding to the experimental results on CIFAR100.
but we can see very similar patterns and make similar observations from the Figures~\ref{Fig:MiniImageNet_1shot} and~\ref{Fig:MiniImageNet_5shot}.

\subsection{Ablation Study}
\label{Sec:Study}

We performed an ablation study 
to verify the effectiveness
of the adversarial training
by comparing CML with the Discriminator and without the Discriminator.
In Table~\ref{AB-study}, we can observe that the Discriminator leads to  $1.75\%$, $0.88\%$, $1.13\%$ and $1.95\%$ improvements in terms of the average accuracy on MiniImageNet and CIFAR100
for 5-way,1-shot and 5-way,5-shot tasks.
As mentioned in Section 4.2, the adversarial training in the meta-update does not have the limitation of few-shot setting due to the relatively more samples for each class.
Meanwhile, 
the Teacher Network, through pre-training on $\mathscr{D}_{meta-train}$ with enough samples, can generate a good feature map for each image, which serves as a learning template for the Student Network.
Thus, the adversarial training, with the help of the pre-trained Teacher Network, can guide the Student Network to obtain discriminative class vector on unseen categories and
achieve a better initialization, from where the tasks, each with a few samples, can be easily handled.

\begin{table}[t]\scriptsize
\centering
\caption{Average accuracies given by CML with (w.) and without (wo) the Discriminator.}
\renewcommand\arraystretch{1.3}
\begin{tabular}{|c|c|c|c|c|}
\toprule
\multirow{2}{*}{Method} & \multicolumn{2}{|c|}{MiniImageNet} & \multicolumn{2}{|c|}{CIFAR100} \\
 & 5-way,1-shot & 5-way,5-shot & 5-way,1-shot & 5-way,5-shot\\ \midrule
CML wo D & $56.91\%$ & $69.6\%$ & $60.95\%$ & $72.08\%$ \\ 
CML w. D& \bm{$58.66\%$} & \bm{$70.48\%$} & \bm{$62.08\%$} & \bm{$74.03\%$} \\ \bottomrule

\end{tabular}
\label{AB-study}
\end{table}

\section{Conclusions}
\label{Sec:Conclusions}

In this paper, we studied a new problem called continual few-shot learning,
in which a sequence of tasks arrives one by one, each of which only contains a few training samples.
We proposed a novel model-agnostic meta-learner, Continual Meta-Learner (CML), which
integrates metric-based classification and a memory-based mechanism along with adversarial
learning into an optimization-based meta-learning framework.
It has been shown by extensive experimental results on MiniImageNet and CIFAR100
that CML consistently outperforms the four baselines in terms of classification accuracy by large margins
in both the 5-way,1-shot and 5-way,5-shot cases, and successfully overcomes catastrophic forgetting; while
the optimization-based meta-learner suffers from severe forgetfulness.

\ifCLASSOPTIONcaptionsoff
  \newpage
\fi



%
\bibliographystyle{IEEEtran}
\bibliography{ContinualMeta-Learning}

%

\begin{IEEEbiography}[{\includegraphics[width=1in,height=1.25in, clip, keepaspectratio]{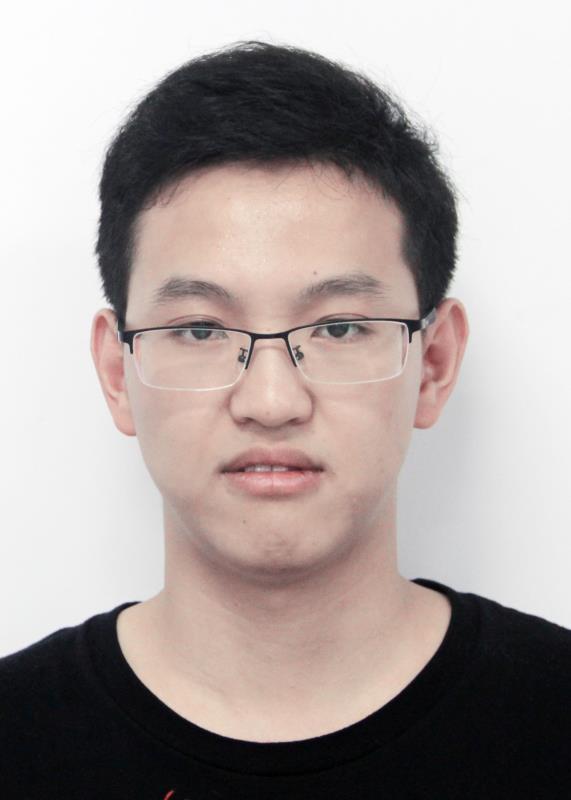}}]{Kun Wu}
received his B.S. degree from Beijing Institute of Technology in 2017. He is currently working towarding his Ph.D degree in the Department of Electrical Engineering and Computer Science at Syracuse University. His research interests include Machine Learning and Computer Vision.
\end{IEEEbiography}

\begin{IEEEbiography}[{\includegraphics[width=1in,height=1.25in, clip, keepaspectratio]{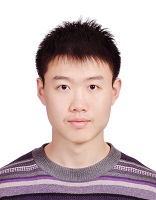}}]{Chengxiang Yin}
received his B.S. degree from Beijing Institute of Technology in 2016. He is currently working toward his Ph.D degree in the Department of Electrical Engineering and Computer Science at Syracuse University. His research interests include Machine Learning and Computer Vision.
\end{IEEEbiography}

\begin{IEEEbiography}[{\includegraphics[width=1in,height=1.25in, clip, keepaspectratio]{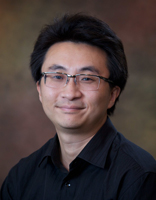}}]{Jian Tang}
(F2019) is a Professor in the Department of Electrical Engineering and Computer Science at Syracuse University and an IEEE Fellow. He received his Ph.D degree in Computer Science from Arizona State University in 2006. His research interests lie in the areas of Machine Learning, IoT, Wireless Networking and Big Data Systems. Dr. Tang has published over 150 papers in premier journals and conferences. He received an NSF CAREER award in 2009. He also received several best paper awards, including the 2019 William R. Bennett Prize and the 2019 TCBD (Technical Committee on Big Data) Best Journal Paper Award from IEEE Communications Society (ComSoc), the 2016 Best Vehicular Electronics Paper Award from IEEE Vehicular Technology Society (VTS), and Best Paper Awards from the 2014 IEEE International Conference on Communications (ICC) and the 2015 IEEE Global Communications Conference (Globecom) respectively. He has served as an editor for several IEEE journals, including IEEE Transactions on Big Data, IEEE Transactions on Mobile Computing, etc. In addition, he served as a TPC co-chair for a few international conferences, including the IEEE/ACM IWQoS'2019, MobiQuitous'2018, IEEE iThings'2015. etc.; as the TPC vice chair for the INFOCOM'2019; and as an area TPC chair for INFOCOM 2017-2018. He is also an IEEE VTS Distinguished Lecturer, the Chair of the Communications Switching and Routing Committee of IEEE ComSoc and an ACM Distinguished Member.
\end{IEEEbiography}

\begin{IEEEbiography}[{\includegraphics[width=1in,height=1.5in,clip,keepaspectratio]{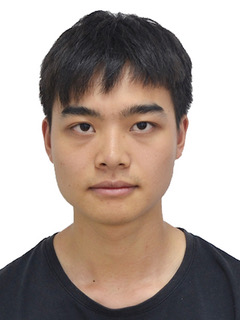}}]{Zhiyuan Xu}
is currently pursuing the Ph.D. degree at the Department of Electrical Engineering and Computer Science, Syracuse University, Syracuse, NY, USA. He received the B.E. degree in School of Computer Science and Engineering from University of Electronic Science and Technology of China, Chengdu, China, in 2015. He was an exchange student in 2013 at Department of Computer Science and Information Engineering, National Taiwan University of Science and Technology, Taipei, Taiwan. He was a visiting student in 2015 at Dalhousie University, Halifax, NS, Canada. His current research interests include Machine Learning and Communication Networks.
\end{IEEEbiography}

\begin{IEEEbiography}[{\includegraphics[width=1in,height=1.25in, clip, keepaspectratio]{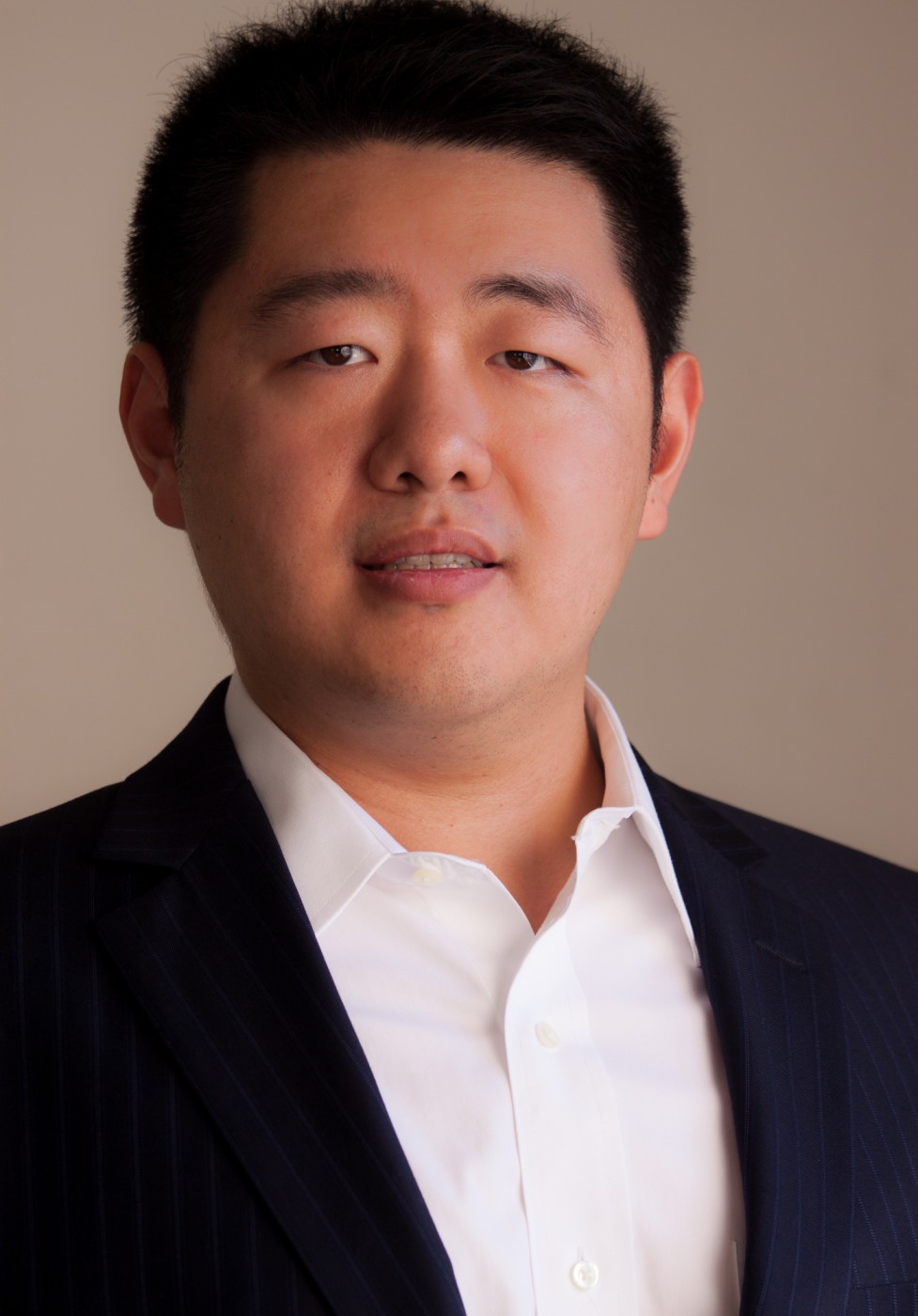}}]{Yanzhi Wang}
is an assistant professor in Department of Electrical and Computer Engineering at Northeastern University.He has received his Ph.D. Degree in Computer Engineering from University of Southern California (USC) in 2014, under supervision of Prof. Massoud Pedram, and his B.S. Degree with Distinction in Electronic Engineering from Tsinghua University in 2009.
Dr. Wang's current research interests are energy-efficient and high-performance implementations of deep learning and artificial intelligence systems. Besides, he works on the application of deep learning and machine intelligence in various mobile and IoT systems, medical systems, and UAVs. His group works on both algorithms and actual implementations (FPGAs, circuit tapeouts including superconducting circuits, mobile and embedded systems, and UAVs). His works have been published in top venues in conferences and journals (e.g. ASPLOS, MICRO, AAAI, ICML, VLDB, FPGA, DAC, ICCAD, DATE, ISLPED, LCTES, INFOCOM, ICDCS, TComputer, TCAD, Plos One, etc.), and have been cited over 4,800 times according to Google Scholar. He has received four Best Paper or Top Paper Awards from major conferences, another eight Best Paper Nominations and two Popular Papers in IEEE TCAD. His group is sponsored by the NSF, DARPA, IARPA, AFRL/AFOSR, Syracuse CASE Center, and industry sources.
\end{IEEEbiography}
 
\begin{IEEEbiography}[{\includegraphics[width=1in,height=1.25in, clip, keepaspectratio]{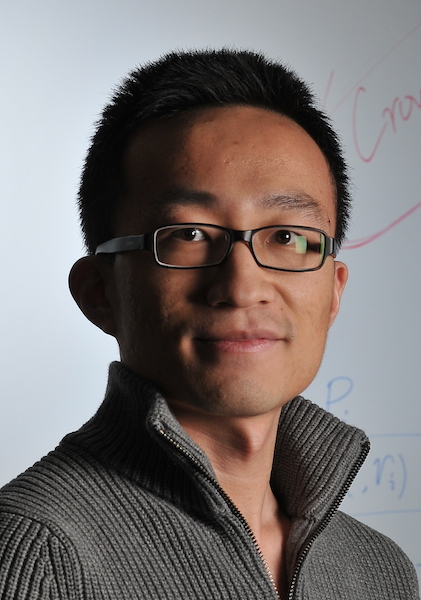}}]{Dejun Yang}
received the B.S. degree in computer science from Peking University, Beijing, China, in 2007, and the Ph.D. degree in computer science from Arizona State University, Tempe, AZ, USA, in 2013. He is currently an Associate Professor of computer science with ColoradoSchool of Mines, Golden, CO, USA. His research interests include Internet of things, networking, and mobile sensing and computing, with a focus on the application of game theory, optimization, algorithm design, and machine learning to resource allocation, security, and privacy problems. He has received the IEEE Communications Society William R. Bennett Prize in 2019 (Best Paper Award for IEEE/ACM TON and IEEE TNSM in the previous three years), and the Best Paper Awards at the IEEE GLOBECOM (2015), the IEEE MASS (2011), and the IEEE ICC (2011 and 2012), as well as the Best Paper Award Runner-up at the IEEE ICNP (2010). He is the TPC Vice Chair of information systems for the IEEE INFOCOM 2020, a Student Travel Grant Co-Chair for INFOCOM 2018–2019, and was the Symposium Co-Chair for the International Conference on Computing, Networking and Communications (ICNC) 2016. He currently serves as an Associate Editor for the IEEE INTERNET OF THINGS JOURNAL.
\end{IEEEbiography}

\end{document}